\documentclass[letterpaper, 10 pt, conference]{ieeeconf}  

\IEEEoverridecommandlockouts                              

\usepackage{graphics} 
\usepackage{epsfig} 
\usepackage{mathptmx} 
\usepackage{times} 
\usepackage{amsmath} 
\usepackage{amssymb}  
\usepackage{graphicx}   
\usepackage{makecell}   
\usepackage{booktabs}   
\usepackage{array}      
\usepackage{tabularx} 
\usepackage{graphicx} 
\usepackage{float} 
\usepackage{subfigure} 
\usepackage{parskip}
\usepackage{multirow}
\usepackage{tcolorbox}
\usepackage{amsmath}
\usepackage{amssymb}
\usepackage{booktabs}
\usepackage{tabularray}
\usepackage{marvosym}

\usepackage[breaklinks,colorlinks]{hyperref}
\makeatletter
\newcommand{\removelatexerror}{\let\@latex@error\@gobble}

\IEEEoverridecommandlockouts
\begin{document}

\title{\LARGE \bf
WcDT: World-centric Diffusion Transformer for Traffic Scene Generation
}
\author{Chen Yang$^{1*}$, Yangfan He$^{2*}$, Aaron Xuxiang Tian$^{3}$, Dong Chen$^{4}$, \\Jianhui Wang\textsuperscript{5}, Tianyu Shi$^{6}$, Arsalan Heydarian$^{7}$, Pei Liu$^{8\textrm{\Letter}}$
\thanks{* Equal contribution.}
\thanks{{\textrm{\Letter}} \text{ Corresponding author.}}
\thanks{$^{1}$Department of Computer Science and Informatics, Cardiff University.
        {\tt\small yc19970530@gmail.com}}%
\thanks{$^{2}$College of Libera Arts, University of Minnesota - Twin Cities.
        {\tt\small he000577@umn.edu}}%
\thanks{$^{3}$Independent researcher, USA.
        {\tt\small aarontian00@gmail.com}}%
\thanks{$^{4}$Agricultural \& Biological Engineering, Mississippi State University.
        {\tt\small dc2528@msstate.edu}}%
\thanks{\textsuperscript{5}Information and Software Engineering, University of Electronic Science and Technology of China. 
        {\tt\small jianhuiwang@std.uestc.edu.cn}}%
\thanks{$^{6}$Transportation Research Institute, University of Toronto.
        {\tt\small ty.shi@mail.utoronto.ca}}%
\thanks{$^{7}$Link Lab \& Civil and Environmental Engineering, University of Virginia.
        {\tt\small heydarian@virginia.edu}}%
\thanks{$^{8}$Intelligent Transportation Thrust, The Hong Kong University of Science and Technology (Guangzhou).
        {\tt\small pliu061@connect.hkust-gz.edu.cn}}
}

\maketitle
\begin{abstract}
In this paper, we introduce a novel approach for autonomous driving trajectory generation by harnessing the complementary strengths of diffusion probabilistic models (a.k.a., diffusion models)  and transformers. Our proposed framework, termed the ``World-centric Diffusion Transformer''(WcDT), optimizes the entire trajectory generation process, from feature extraction to model inference. To enhance the scene diversity and stochasticity, the historical trajectory data is first preprocessed into ``Agent Move Statement'' and encoded into latent space using Denoising Diffusion Probabilistic Models (DDPM) enhanced with Diffusion with Transformer (DiT) blocks. Then, the latent features, historical trajectories, HD map features, and historical traffic signal information are fused with various transformer-based encoders that is used to enhance the interaction of agents with other elements in the traffic scene. The encoded traffic scenes are then decoded by a trajectory decoder to generate multimodal future trajectories. Comprehensive experimental results show that the proposed approach exhibits superior performance in generating both realistic and diverse trajectories, showing its potential for integration into automatic driving simulation systems. Our code is available at \url{https://github.com/yangchen1997/WcDT}.
\end{abstract}
\begin{figure}[!ht]
  \centering
  \includegraphics[width=\linewidth]{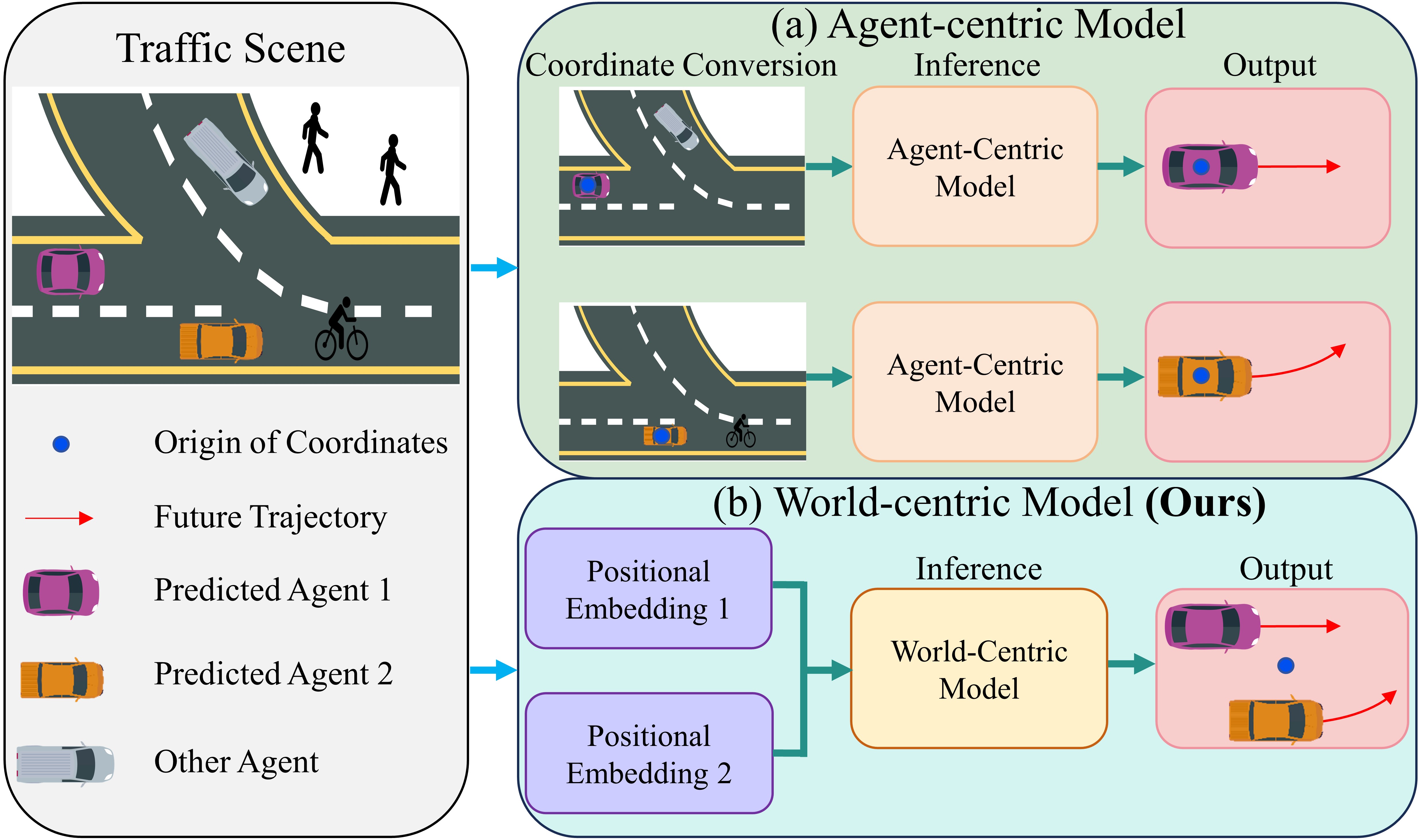}
  \vspace{1pt}
  \caption{
        World-centric model and agent-centric model: (a) The conventional "Agent-centric model" is common in trajectory prediction, including on the Sim Agents leaderboard. (b) Our approach replaces complex coordinate transformations with position embeddings, enhancing efficiency in multi-scenario, multi-agent trajectory generation.
  }
  \label{fig:difference_gene_traj_and_pred}
  \vspace{-0.5cm}
\end{figure}
\section{Introduction}
\label{sec:intro}
Autonomous driving is a transformative technology aimed at reducing driver fatigue and traffic congestion by enabling autonomous vehicle operation \cite{huang2022survey, teng2023motion, paden2016survey, chen2024communication}. Developing these algorithms involves iterative optimization for safety and performance \cite{chen2022milestones, liu2023systematic, ge2023use}, but real-world testing poses challenges due to time constraints, safety concerns, regulatory hurdles, and high costs \cite{o2018scalable, hu2023simulation}. Simulators play a vital role in the cost-effective testing and evaluation of autonomous driving systems (ADS) by providing controllable environments \cite{grollius2021concept, weiss2022high}. To be effective, they must realistically replicate traffic scenarios and driver behaviors \cite{zhong2023guided}. Current simulators rely on replaying driving logs or heuristic controllers \cite{treiber2000congested, zhong2023guided}, limiting diversity and unpredictability in real-world behavior, which impacts ADS validation \cite{guo2023scenedm}. Multimodal motion prediction approaches \cite{chai2019multipath, varadarajan2022multipath++, ngiam2021scene, suo2021trafficsim, shi2022motion, wang2023multiverse} have shown promise in traffic scene generation but struggle to generate diverse actions for all agents using comprehensive global information \cite{chai2019multipath}. Generative adversarial networks (GANs) and Variational Auto-Encoders (VAEs) have been applied to traffic scene generation but face limitations. These models often lack diversity, reflecting training data distributions \cite{zhong2023guided}, and GANs suffer from unstable adversarial training \cite{guo2023scenedm}. Additionally, they fail to capture agent trajectory smoothness, leading to unrealistic results \cite{guo2023scenedm}, and typically focus on individual vehicle paths, neglecting all agents. Recently, diffusion models have emerged as a promising alternative for diverse traffic scenarios \cite{janner2022planning, zhong2023guided, guo2023scenedm, niedoba2023diffusion}, treating generation as an inverse diffusion process. However, these models often require agent-centric Cartesian coordinates \cite{chai2019multipath} and generate only one trajectory per agent per inference. In this paper, we propose a novel framework for traffic scene generation tailored to autonomous driving, leveraging diffusion models and transformer-based encoder-decoder architectures. Our "World-centric Diffusion Transformer" (WcDT) framework optimizes trajectory generation from feature extraction to inference, enabling coherent and joint future movements for various agents in a single inference. Our contributions include:
\begin{itemize}
\item A new paradigm for simultaneous, consistent future movement generation for all agents in a single inference.
\item A Diffusion-Transformer module that enhances scene diversity and stochasticity, integrating the world state efficiently.
\item Benchmarking performance for realism and diversity in trajectory generation, validated on open traffic datasets.
\end{itemize}
\section{Related work}\label{sec:2}
\subsection{Motion prediction-based methods}
Recent developments in traffic scene generation utilize motion prediction methodologies to enhance realism in multimodal scenarios \cite{chai2019multipath, varadarajan2022multipath++, ngiam2021scene, suo2021trafficsim, shi2022motion, wang2023multiverse}. For example, Multipath++ \cite{varadarajan2022multipath++} advances its predecessor by integrating a context-aware fusion approach with Gaussian mixture models for more precise trajectory predictions. Similarly, Trafficsim \cite{suo2021trafficsim} employs an implicit latent variable model for simulating multi-agent interactions. Transformer-based encoder-decoder architectures are also central to motion prediction \cite{ngiam2021scene, shi2022motion, wang2023multiverse}. The Scene Transformer \cite{ngiam2021scene} encodes interactions among agents using a global coordinate frame, enabling joint behavior prediction. The Motion Transformer (MTR) \cite{shi2022motion} optimizes both global intention and local movement, achieving top rankings in the Waymo Open Motion Dataset \cite{sun2021large}. Leading the Waymo Open Sim Agents Challenge (WOSAC), the Multiverse Transformer (MVTA) introduces novel training and sampling methods along with a receding horizon prediction technique. However, while these approaches focus on local scene details, they often overlook the broader multimodal context. Our diffusion-based model addresses this by generating diverse actions for all agents in each inference, overcoming a significant limitation of traditional trajectory prediction methods.
\subsection{Generative model-based methods}
Generative adversarial networks (GANs) \cite{li2019conditional, bhattacharyya2022modeling} and Variational Auto-Encoders (VAEs) \cite{ding2019multi, oh2022cvae} have been utilized for traffic scene generation. For example, \cite{li2019conditional} proposes a conditional generative neural system (CGNS) for probabilistic trajectory generation, while \cite{oh2022cvae} develops a conditional VAE for multimodal, context-driven traffic scene generation. However, these methods often generate unrealistic trajectories due to their reliance on training data distribution and limited diversity \cite{zhong2023guided}. Additionally, GANs can suffer from unstable training \cite{guo2023scenedm}, and VAEs may be constrained by a simple Gaussian prior, limiting their expressiveness. Recently, diffusion models have emerged as a promising alternative to GANs and VAEs for generating realistic and diverse data \cite{janner2022planning, zhong2023guided}. Notably, \cite{janner2022planning} applies a classifier-guided diffusion approach to trajectory data with a probabilistic framework, while \cite{zhong2023guided} introduces a conditional diffusion model for controllable traffic generation (CTG), allowing users to specify desired trajectory properties while maintaining realism and physical plausibility. However, these methods primarily focus on single-agent behaviors. Recent work on multi-agent trajectory generation using diffusion models includes SceneDM \cite{guo2023scenedm}, which generates future motions of all agents, achieving state-of-the-art results on the Waymo Sim Agents Benchmark, and DJINN \cite{niedoba2023diffusion}, which produces traffic scenarios based on the joint states of all agents. A limitation of these models is that they predict individual agent trajectories per inference. In contrast, our approach integrates diffusion models with transformer-based encoder-decoder architectures to simultaneously generate joint, coherent future trajectories for all agents.
\begin{figure*}[!ht]
    \centering
    \includegraphics[width=1.0\textwidth]{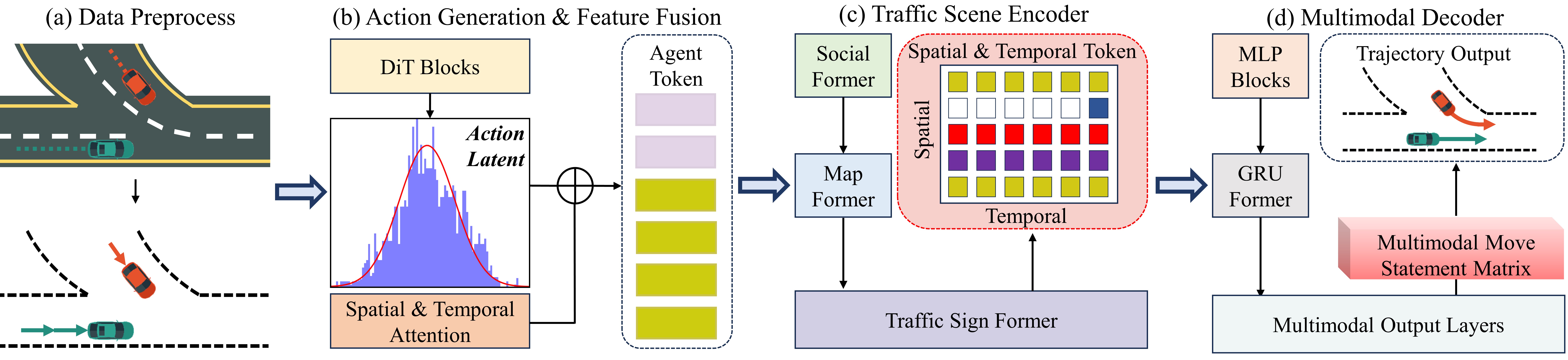}
    \caption{
        Overview of WcDT, which consists of the following modules: (a) Agent action generation and agent to agent cross attention blocks; (b) The traffic scene encoder extracts temporal and spatial features in the traffic scene, including: other agents, traffic signals, HD maps; (c) The multimodal trajectory decoder is used to generate possible future actions for all predicted agents. 
    }
    \label{fig:encoder}
    \vspace{-0.5cm}
\end{figure*}

\section{Methodology}\label{sec:3}
In this section, we present our novel WcDT framework for representing and generating complex traffic scenes. We first explain how traffic environments are modeled, followed by an introduction to the framework and its components.
\subsection{Traffic scene representation}
Traffic environments are composed of multimodal data, such as road layouts, traffic signals, agent movements, and environmental conditions \cite{nayakanti2023wayformer, gao2020vectornet}. To encode these elements in WcDT, we adopt a unified approach that captures both predicted and environmental (world) agents. Unlike existing methods that require transforming information to each agent’s perspective \cite{chai2019multipath}, our approach simplifies this by:
\begin{itemize}
    \item Using a unified Cartesian coordinates system for both predicted and world agents.
    \item Representing historical agent trajectories through movement statements instead of traditional coordinate vectors.
\end{itemize}
Key variables for simulating traffic scenarios in WcDT:
\begin{itemize}
    \item $\mathcal{A}_{all}$, $\mathcal{A}_p$, $\mathcal{A}_w$: Counts of all agents, predicted agents, and world agents, respectively.
    \item $\mathcal{T}_h$, $\mathcal{T}_f$: Historical and future time steps.
    \item $\mathcal{L}$, $\mathcal{P}$: Lane lines and points within scenarios.
    \item $\mathcal{S}_{tl}$: Traffic light states.
    \item $\mathcal{D}$: Dimensionality of different traffic elements in a traffic scenario ($\mathcal{D}_a$ represents the features of an agent, $\mathcal{D}_t$ represents the features of a traffic light, and $\mathcal{D}_m$ represents the features of a map element).
\end{itemize}
For different traffic objects, we represent them as follows:
\begin{itemize}
    \item \textit{Agent move statement and features:} To mitigate the impact of varying agent positions on historical and future trajectories, we introduce absolute states for past and prospective agent states. For agent $i$ at time step $t$, the state $s_t^i$ is defined as $s_t^i=[(x_t-x_{t-1}), (y_t-y_{t-1}), (\theta_t-\theta_{t-1}), (v_t-v_{t-1})]$, where $x_t$, $y_t$, $\theta_t$, and $v_t$ represent longitudinal position, lateral position, heading angle, and velocity, respectively. The feature space for each agent is $[\mathcal{A}, \mathcal{T}_h-1, \mathcal{D}_{a}]$.
    \item \textit{Traffic light feature:} The traffic light dataset for each scenario, denoted as $[\mathcal{S}_{tl}, \mathcal{T}_{h}, \mathcal{D}_{t}]$, contains the positions and operational statuses of signals over historical intervals. For any traffic signal point $s_{tl} \in \mathcal{S}_{tl}$, this information is represented using a one-hot encoding of signal states and spatial positions at each historical moment.
    \item \textit{Map feature:} The map features, denoted as $[1, \mathcal{L}, \mathcal{P}, \mathcal{D}_m]$, encompass key lane details in a traffic scenario, including positions and types. Each lane line $l_t \in \mathcal{L}$ at the current time step is represented positions of all points along the lane and using one-hot encoding to specify the its type.
\end{itemize}
Figure~\ref{fig:encoder} shows an overview of our proposed WcDT framework for traffic scene generation, including three major components: action diffusion, scene encoder, and trajectory decoder, which are detailed in the following subsections.
\subsection{Action Diffusion}
\begin{figure}[!ht]
  \centering
  \includegraphics[width=\linewidth]{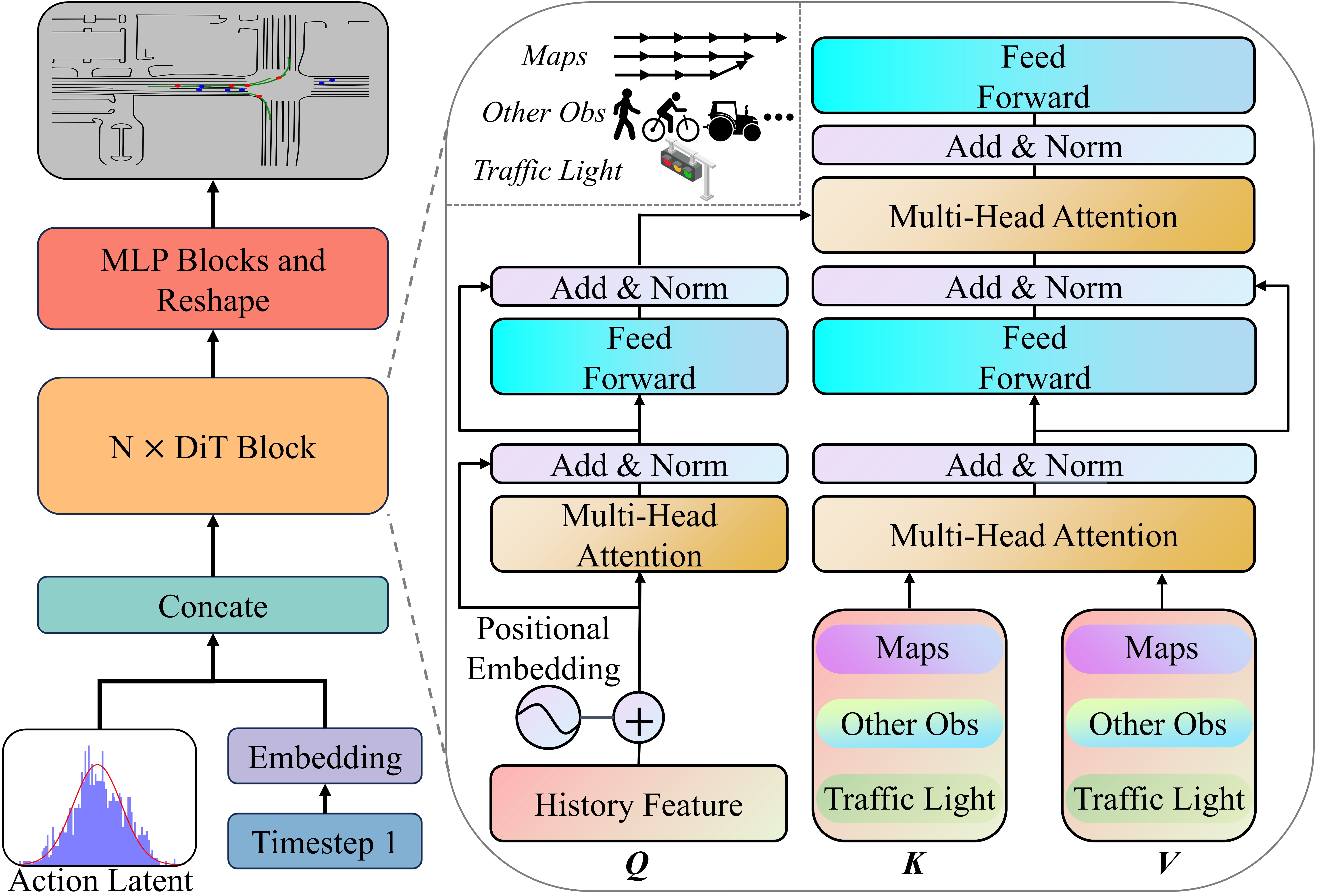} 
  \caption{Overview of the developed conditional DiT blocks and illustration of DiT block processing, where action latent features are integrated with map, object, and traffic light using multi-head attention for traffic scene generation.}
  \label{fig:dits}
  \vspace{-0.4cm}
\end{figure}

To enhance trajectory diversity in WcDT, we encode agent actions into latent space to increase variability. These latent features are then input into the scene encoder. We use Denoising Diffusion Probabilistic Models (DDPM) \cite{ho2020denoising} for action encoding. Although DDPM traditionally employs U-Net architectures, recent research \cite{peebles2023scalable} demonstrates that transformers can achieve comparable performance without U-Net’s inductive biases. Consequently, we replace U-Net with Diffusion Transformers (DiTs) to improve performance and ensure diverse agent trajectories.  Figure~\ref{fig:dits} shows the architecture of conditional DiT blocks for encoding "latent action features". The network takes random noise, time steps, and historical trajectories as inputs and produces latent action features for the scene encoder. The DDPM loss function guides the network to generate latent features consistent with agent kinematics \cite{ho2020denoising}, thus enhancing trajectory variability. The loss function for the DiT module is as follows:
\begin{align}
\centering
\mathcal{L}_{diff} = ||\epsilon-\epsilon_{\theta}(
\sqrt{\Bar{\alpha_t}}x_{0} + \sqrt{1-\Bar{\alpha_t}}\epsilon, t
)||^2,
\label{eq:loss_dfff}
\end{align}
where $\Bar{\alpha_t}$ are hyperparameters for diffusion model training, $\epsilon_{\theta}$ represents the diffusion model with DiT blocks, and $\epsilon$ is Gaussian noise.
\subsection{Scene Encoder}
In traffic scenes, agents like vehicles, pedestrians, and bicycles, along with map features and traffic signals, are present. To generate diverse trajectories, we use embedding blocks of different sizes and layers. These blocks encode agents' characteristics, bypassing the need for agent-specific coordinate transformations. The Pose-Embedding encodes positional data $p_i$ into a 1D matrix, while Feature-Embedding translates attributes like height, width, and type into another matrix. For agent $i$:
\begin{align}
E_p & = \phi_{p}[x_i, y_i], \quad E_f = \phi_{f}[f_w, f_h, f_{type}],
\end{align}
where $\phi_p$ and $\phi_f$ represent linear transformations. The final agent embedding $E_A$ is:
\begin{align}\label{eq:al}
E_{A} = \text{ReLU}(\text{LayerNorm}(\text{Concat}(E_p, E_f))).
\end{align}
To represent traffic scenarios, the encoding process integrates world agents, maps, and traffic lights. Features are processed into embeddings, refined via neural network blocks, including multi-head self-attention for detailed analysis and cross-attention for feature relationships. Attention layers, replacing CNNs, dynamically capture long-range dependencies. The self-attention encoding for world agents is:
\begin{align}
q_{A_p} = W^{Q \times h} E_p, \quad k_{A_p} = W^{K \times h} E_p, \quad v_{A_p} = W^{V \times h} E_p,
\end{align}
where $W^Q$, $W^K$, $W^V$ are learnable parameters, and attention is calculated as:
\begin{align}
\propto_{A_p} = \text{Softmax}\left(\frac{q_{A_p}^T}{\sqrt{d_k}} k_{A_p}\right), \quad \text{Self}_{A_p} = \propto_{A_p} v_{A_p}.
\end{align}
Cross-attention for encoding map and traffic light features follows a similar process:
\begin{align}
q_{A_p} = W^{Q \times h} E_{A_p}, \quad k_m = W^{K \times h} E_m, \quad v_m = W^{V \times h} E_m, \\
\propto_m = \text{Softmax}\left(\frac{q_{A_p}^T}{\sqrt{d_k}} k_M\right), \quad \text{Cross}_m = \propto_{A_p} v_M.
\end{align}
\textbf{Spatial and Temporal Fusion.} We propose Temporal-Spatial Fusion Attention layers to capture the dynamic nature of traffic scenarios by integrating multimodal data. The agent's features are augmented with latent action features (from Eq.~\ref{eq:al}) and processed through self-attention layers to identify key temporal-spatial insights, ensuring an accurate understanding of traffic dynamics.
\subsection{Trajectory Decoder}
\begin{figure}[t!]
    \centering
    \includegraphics[width=0.45\textwidth]{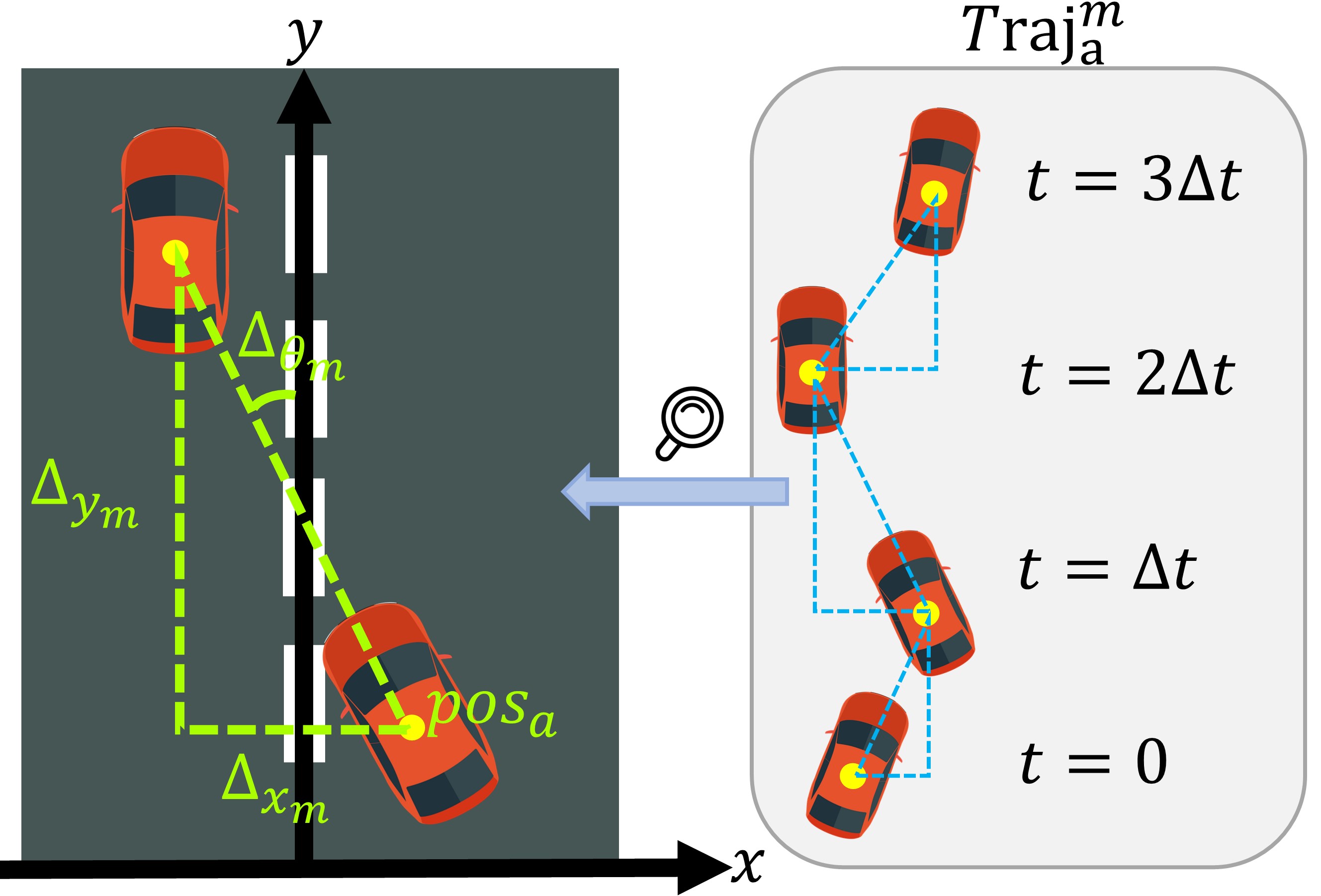}
    \caption{Illustration of the trajectory generation process.}
    \label{fig:trajectory_decoder}
    \vspace{-0.3cm}
\end{figure}

The trajectory decoder translates fused traffic features into agents' future trajectories using GRU and MLP blocks. Drawing inspiration from \cite{chai2019multipath}, we employ a multimodal output mechanism to handle agents with varied behaviors. To reduce the influence of different initial positions, the decoder outputs agents' move statements and their likelihoods. The trajectory for model $\mathcal{M}$ is computed as shown in Fig.~\ref{fig:trajectory_decoder}:
\begin{align}
\centering
\text{Traj}^{m}_{a} = \text{Pos}_{a} + \sum_{i=t}^{T_f} [\Delta x_{m}, \Delta y_{m}, \Delta \theta_{m}],
\end{align}
where $\text{Pos}_{a}$ is the agent's current position, and $\text{Traj}^{m}_{a}$ denotes future trajectory points, computed by adding the displacement $[\Delta x_{m}, \Delta y_{m}, \Delta \theta_{m}]$ for each time step. The speed is:
\begin{align}
\text{Speed}^{m}_a = \frac{[\Delta x_m, \Delta y_m]}{\Delta t},
\end{align}
calculated from the displacement over time $\Delta t$. This kinematic approach outputs the trajectory as $[x^m_a, y^m_a, \theta^m_a, v^m_a]$.
\subsection{Loss functions}
Our model aims to ensure generated trajectories adhere to scene constraints while maintaining diversity. The trajectory with the lowest loss from the multimodal set is selected, and its deviation from the ground truth is measured using the Huber loss \cite{2001Greedy}:
\begin{align}
\centering
\mathcal{L}_{reg} = \text{Huber}(\text{Traj}_{p}, \text{Traj}_{gt}),
\label{eq:reg}
\end{align}
where $\text{Traj}_{p}$ and $\text{Traj}_{gt}$ are the predicted and ground truth trajectories, respectively. We also introduce a classification loss to identify the modality closest to the ground truth, where the modality with the smallest AED is used as the classification target:
\begin{align}
\centering
\mathcal{L}_{cls} = -\sum_{i=1}^{M}y_{i}log(p_{i}),
\end{align}
The total loss is a combination of diffusion, regression, and classification losses:
\begin{align}
\centering
\mathcal{L}_{total} = \mathcal{L}_{diff} + \mathcal{L}_{reg} + \mathcal{L}_{cls}.
\end{align}
Here, $\mathcal{L}_{diff}$ is the standard diffusion model loss, computed as the L2 loss between predicted and original noise.

\begin{table*}[!ht]
    \centering
    \caption{The Sim Agents Leaderboard results evaluate methods using 10 similarity metrics (kinematic, object interaction, and map-based) and distance error metrics (ADE and MinADE), where higher similarity values indicate better performance, and lower ADE/MinADE values represent more accurate trajectory predictions.}
    \resizebox{0.98\textwidth}{!}
    {\begin{tabular}{cccccccccccc|cc}
        \toprule
        & \textbf{\makecell[c]{Method}} & \textbf{\makecell[c]{Linear\\Speed}} & \textbf{\makecell[c]{Linear\\Accel}} 
        & \textbf{\makecell[c]{Ang\\Speed}} & \textbf{\makecell[c]{Ang\\Accel}} & \textbf{\makecell[c]{Dist\\to Obj}} 
        & \textbf{\makecell[c]{Collision}} & \textbf{\makecell[c]{TTC}} & \textbf{\makecell[c]{Dist to\\Road Edge}} 
        & \textbf{\makecell[c]{Offroad}} & \textbf{\makecell[c]{Composite\\Metric}} & \textbf{\makecell[c]{ADE}} 
        & \textbf{\makecell[c]{MinADE}} \\
        \hline \hline
        & \textbf{\makecell[c]{Random Agent~\cite{montali2023waymo}}} & 0.002 & 0.044 & 0.074 & 0.120 & 0.000 & 0.006 
        & 0.734 & 0.178 & 0.325 & 0.163 & 50.740 & 50.707\\
        & \textbf{\makecell[c]{Constant Velocity~\cite{montali2023waymo}}} & 0.074 & 0.058 & 0.019 & 0.035 & 0.208 & 0.202 
        & 0.737 & 0.454 & 0.325 & 0.238 & 7.924 & 7.924\\
        & \textbf{\makecell[c]{MTR+++~\cite{shi2023mtr++}}} & 0.412 & 0.107 & 0.484 & 0.437 & 0.346 & 0.414 & 0.797 
        & 0.654 & 0.577 & 0.470 & 2.129 & 1.682\\
        & \textbf{\makecell[c]{WayFormer~\cite{nayakanti2023wayformer}}} & 0.408 & 0.127 & 0.473 & 0.437 & 0.358 
        & 0.403 & 0.810 & 0.645 & 0.589 & 0.472 & 2.588 & 1.694\\
        & \textbf{\makecell[c]{MULTIPATH++~\cite{varadarajan2022multipath++}}} & 0.432 & 0.230 & 0.515 & 0.452 & 0.344 
        & 0.420 & 0.813 & 0.639 & 0.583 & 0.489 & 5.308 & 2.052\\
        & \textbf{\makecell[c]{MVTA~\cite{wang2023multiverse}}} & 0.437 & 0.220 & 0.533 & 0.481 & 0.373 & 0.436 
        & 0.830 & 0.654 & 0.629 & 0.509 & 3.938 & 1.870\\
        & \textbf{\makecell[c]{MVTE~\cite{wang2023multiverse}}} & 0.443 & 0.222 & 0.535
        & 0.481 & 0.382 & 0.451 & 0.832 & 0.664 & \textbf{0.641} & 0.517 & 3.873 & 1.677\\
        & \textbf{WcDT (Ours)} & \textbf{0.515} & \textbf{0.370} & \textbf{0.543} & \textbf{0.508} & \textbf{0.548} & \textbf{0.629} & \textbf{0.846} & \textbf{0.738} & 0.608 & \textbf{0.743} & \textbf{2.045} & \textbf{1.472}\\
         \bottomrule
    \end{tabular}}
    \label{tab:comparison results}
    \vspace{-0.5cm}
\end{table*}
\section{Experiments and Results} \label{sec:4}
\subsection{Experimental Setup}
\textbf{Dataset.} We use the Waymo Motion Prediction dataset \cite{sun2021large}, containing 576,012 driving scenarios. The data is divided into 486,995 training, 44,097 validation, and 44,920 testing scenarios. Each scenario lasts 9 seconds, sampled at 10 Hz, with only the first second of the testing scenarios available for generating future trajectories for the next 8 seconds.
\textbf{Metrics.} We employ established evaluation metrics \cite{montali2023waymo,gil1991fast,suo2021trafficsim} and Sim Agents Challenge metrics \cite{montali2023waymo} to assess the realism and diversity of generated trajectories. These metrics cover kinematic, object interaction, and map-based aspects. We minimize the negative log-likelihood (NLL):
\begin{align}
NLL^{*} = -\frac{1}{|\mathcal{D}|} \sum_{i=0}^{|\mathcal{D}|} 
Logq^{world}(o_{\geq t,i}|o_{<t,i}),
\end{align}
where $o_{<t,i}$ represents historical observations, and $o_{\geq t,i}$ denotes future observations.\\
\textbf{Implementation Details.} We train our model for 128 epochs using two NVIDIA A100 GPUs, with Adam optimizer \cite{2014Adam}. We set the batch size to 128, the initial learning rate to $2\times10^{-4}$, and apply a cosine annealing scheduler \cite{2016SGDR} for learning rate adjustment. The architecture includes 2 DiT blocks, 4 Other Agent Former blocks, 4 Map Former blocks, and 2 Traffic Light Former blocks, with the Trajectory Decoder using 2 MLP blocks. Both Multi-Head Self-Attention and Cross-Attention mechanisms are configured with 8 attention heads. We test two model variants: WcDT-64 (64 hidden units) and WcDT-128 (128 hidden units). The origin for all scenarios is set at the current self-driving vehicle's location.

\subsection{Comparison with state-of-the-art methods} 
We compare our proposed WcDT model against several state-of-the-art benchmarks submitted to the Sim Agent Challenge\footnote{Sim Agent Leaderboard as of 02-04-2024: \url{https://waymo.com/open/challenges/2023/sim-agents/}}~\cite{montali2023waymo}, including Random Agent, Constant Velocity~\cite{montali2023waymo}, MTR+++~\cite{shi2023mtr++}, WayFormer~\cite{nayakanti2023wayformer}, MULTIPATH++~\cite{varadarajan2022multipath++}, MVTA~\cite{wang2023multiverse}, and MVTE~\cite{wang2023multiverse}. MVTE, MVTA, and MTR+++ show advanced capabilities in generating realistic and feasible motion trajectories for autonomous vehicles. Table~\ref{tab:comparison results} summarizes the evaluation results. The Random Agent method~\cite{montali2023waymo}, which generates random trajectories, performs the worst with a composite score of 0.163. Constant Velocity~\cite{montali2023waymo}, which predicts based on the last known heading and speed, improves slightly, scoring 0.238 in the composite metric. Our WcDT model achieves the highest composite metric of 0.743, indicating strong performance and outperforming MVTE~\cite{wang2023multiverse} in specific metrics such as Linear Speed, Linear Acceleration, Angle Speed, Distance to Object, and Distance to Road Edge, while also achieving a better MinADE score. This highlights WcDT's ability to generate precise and contextually appropriate trajectory predictions, showing its strength in interpreting dynamic traffic environments.
\subsection{Ablation studies on diffusion model}
We evaluate the effect of the "latent action features" encoding, comparing random noise inputs, the Unet network, and our custom DiT block. As shown in Table~\ref{table:ablation_diffsuion}, the DiT module consistently achieves the lowest ADE and MinADE scores, along with the highest composite score. This demonstrates that the diffusion model with the DiT block significantly improves action diversity while maintaining realistic trajectories.
\begin{table}[!ht]
  \centering
  \caption{Ablation Study on Diffusion Model: Evaluating the diffusion model's contribution and comparing impacts of Dit Blocks and Unet.}
  \scalebox{0.8}{
    \begin{tabular}{cccc|ccc}
      \toprule
      \textbf{\makecell[c]{Random Noise}} & 
      \textbf{\makecell[c]{Unet}} & 
      \textbf{\makecell[c]{Dit Blocks}} &  &
      \textbf{\makecell[c]{ADE}}\ensuremath{\downarrow} & 
      \textbf{\makecell[c]{minADE}}\ensuremath{\downarrow} &
      \textbf{\makecell[c]{Composite\\Metric}}\ensuremath{\uparrow} \\
      \midrule  \midrule
      \checkmark &            &            &  & 4.843 & 2.715 & 0.326 \\
                 & \checkmark &            &  & 4.163 & 1.907 & 0.480 \\
                 &            & \checkmark &  & \textbf{2.045} & \textbf{1.472} & \textbf{0.743} \\
      \bottomrule
    \end{tabular}
    }
\label{table:ablation_diffsuion}
\vspace{-0.4cm}
\end{table}
\begin{table}[!ht]
  \centering
   \caption{Ablation Study on Scene-Encoder Components: We assess the importance of each module by adding or removing it from the scene encoder and evaluating performance using ADE and minADE.}
  \scalebox{0.78}{
    \begin{tabular}{ccccc|cc}
      \toprule
      \textbf{\makecell[c]{Spatial \&Temporal\\Attention}} & 
      \textbf{\makecell[c]{Other Agent \\Former}} & 
      \textbf{\makecell[c]{HD Map \\Former}} & 
      \textbf{\makecell[c]{Traffic Light \\Former}} &  &
      \textbf{ADE}\ensuremath{\downarrow} & \textbf{minADE}\ensuremath{\downarrow} \\
      \midrule  \midrule
        & \checkmark & \checkmark & \checkmark &  & 3.035 & 1.973 \\
        \checkmark &            & \checkmark & \checkmark &  & 3.490 & 1.883 \\
      \checkmark & \checkmark &            & \checkmark &  & 2.960 & 2.130 \\
      \checkmark & \checkmark & \checkmark &            &  & 2.593 & 1.865 \\
       
      \checkmark & \checkmark & \checkmark & \checkmark &  & \textbf{2.045} & \textbf{1.472} \\
      \bottomrule
    \end{tabular}
    }
\label{table:ablation_encoder}
\vspace{-0.3cm}
\end{table}
\subsection{Ablation studies on traffic scene encoder}
As shown in Table~\ref{table:ablation_encoder}, the HD Map Former is crucial for accurate trajectory generation, with its removal leading to the worst ADE and MinADE scores. The Spatial and Temporal Attention blocks significantly enhance the encoder’s understanding of traffic, with their absence resulting in sub-optimal ADE (3.035) and MinADE (1.973) scores. The Traffic Light and Other Agent Formers further boost accuracy, and the full encoder setup delivers the best results, demonstrating the effectiveness of our approach.
\begin{figure*}[!ht]
    \centering
    \includegraphics[width=\linewidth]{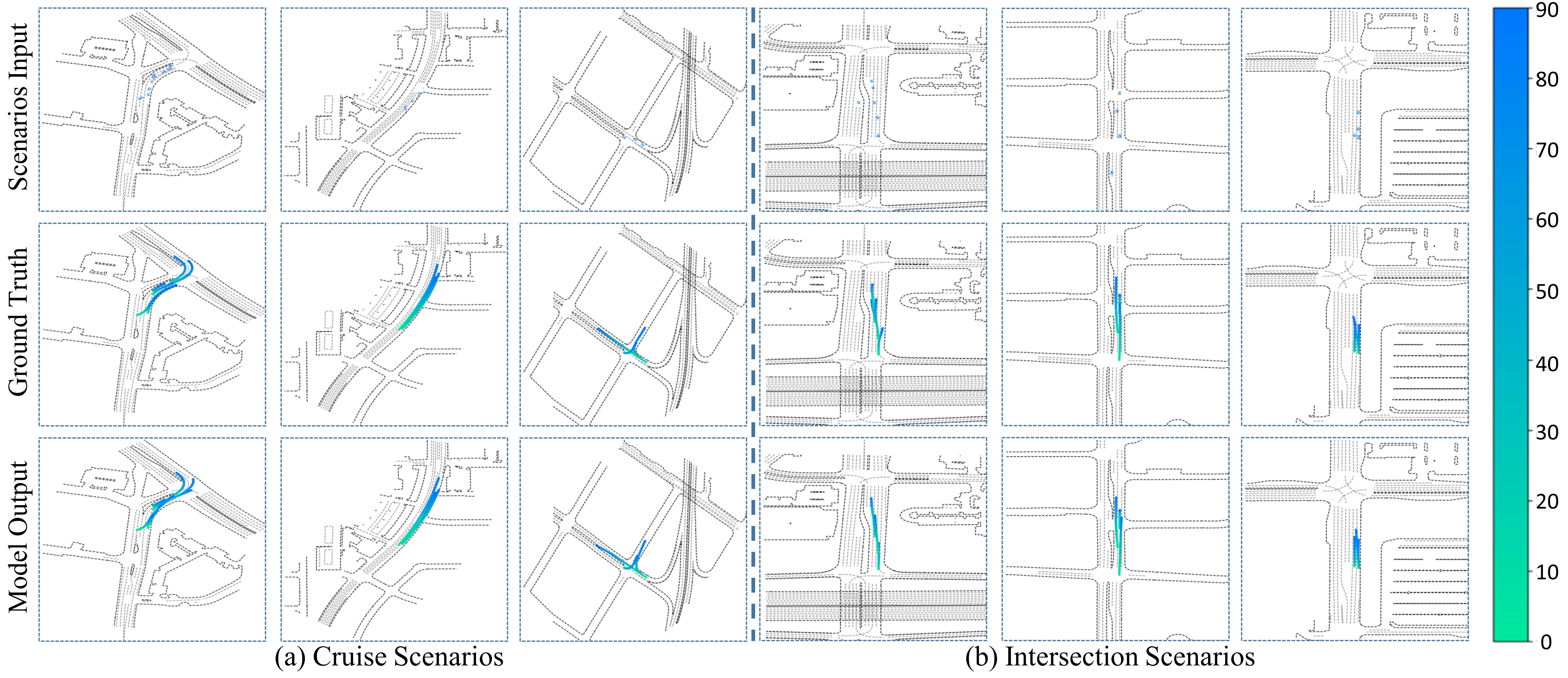}
    \caption{
        Visualization results of the ground truth and WTSGM-generated trajectories in cruise scenarios.
    }
    \label{fig:demo}
\end{figure*}
\subsection{Ablation studies on trajectory decoder} 
We evaluate the trajectory decoder by assessing the contributions of GRU and MLP blocks, focusing on how different configurations impact trajectory prediction. As shown in Table~\ref{table:ablation-decoder}, MLP layers play a crucial role, with their absence resulting in the highest ADE and MinADE scores (3.759 and 2.475). This highlights the importance of MLP blocks in refining the decoder's predictive capabilities. GRU blocks are also essential, helping the model leverage historical data effectively. The best results are achieved when both MLP and GRU blocks are used, demonstrating their combined impact on enhancing trajectory generation.
\begin{table}[!ht]
  \centering
  \caption{Ablation Study on the Trajectory Decoder: We assessed the impact of various components and the dimensionality of hidden units in the trajectory decoder on ADE and MinADE performance.}
  \scalebox{0.9}{
    \begin{tabular}{cccc|cc}
      \toprule
      \textbf{\makecell[c]{MLP Blocks}} & 
      \textbf{\makecell[c]{GRU Blocks}} & 
      \textbf{\makecell[c]{Dimension}} & &
      \textbf{ADE}\ensuremath{\downarrow} & 
      \textbf{minADE}\ensuremath{\downarrow} \\
      \midrule \midrule
     & \checkmark & 64  & & 4.174 & 2.532 \\
     & \checkmark & 128 & & 3.759 & 2.475 \\
      \checkmark &            & 64  & & 3.612 & 1.805 \\
      \checkmark &            & 128 & & 3.857 & 1.780 \\
      
      \checkmark & \checkmark & 64  & & 2.857 & 1.735 \\
      \checkmark & \checkmark & 128 & & \textbf{2.045} & \textbf{1.472} \\
      \bottomrule
    \end{tabular}
    }
\label{table:ablation-decoder}
\vspace{-0.3cm}
\end{table}
\subsection{Ablation studies on network structures} 
We examine the effects of varying the number of modalities and attention heads on trajectory generation. Table~\ref{table:robustness} shows that WcDT-128 consistently outperforms WcDT-64, indicating that more attention layers enhance prediction accuracy. Additionally, the multi-modality configuration in WcDT-64 yields better results than the single-modality setup, as it enables the model to process more information, improving its understanding of the driving environment and leading to more precise predictions.
\begin{table}[!ht]
  \centering
  \caption{Impact of multimodal trajectory decoders and dimension of attention blocks on scene generation performance}
  \scalebox{0.9}{
    \begin{tabular}{cc|cc|cc|cc}
      \toprule
      \textbf{\makecell[c]{Method}} & & 
      \textbf{\makecell[c]{Multimodal}} & &
      \textbf{\makecell[c]{Attention \\Block Heads}} & &
      \textbf{ADE}\ensuremath{\downarrow} & \textbf{minADE}\ensuremath{\downarrow} \\
      \midrule \midrule
      \makecell[c]{WcDT-64} & & 1 & & 8 & & 3.758 & 3.758 \\
      \makecell[c]{WcDT-64} & & 10 & & 8 & & 3.475 & 2.548 \\
      \makecell[c]{WcDT-64} & & 10 & & 16 & & 3.729 & 2.470 \\
      \makecell[c]{WcDT-64} & & 30 & & 16 & & 3.647 & 1.962 \\
      \makecell[c]{WcDT-128} & & 10 & & 8 & & 2.948 & 1.781 \\
      \makecell[c]{WcDT-128} & & 30 & & 16 & & \textbf{2.045} & \textbf{1.472} \\
      \bottomrule
    \end{tabular}
    }
\label{table:robustness}
\vspace{-0.5cm}
\end{table}
\subsection{Visualization Results}
Figure~\ref{fig:demo} illustrates the generated trajectories for randomly sampled Waymo dataset scenarios. The input features include map elements (black dotted lines) and the initial 1-second trajectories of various agents (dots), with each agent’s trajectory represented by a unique color. The intensity of the colors deepens over time, reflecting the temporal progression of each agent's movement. The left side demonstrates lane-changing maneuvers, highlighting the model's ability to predict diverse and accurate trajectories in dynamic driving conditions. On the right, the figure showcases the model’s performance in more complex intersection scenarios, further underscoring its robustness and precision in handling challenging traffic environments.
\section{Conclusion}\label{sec:5}
This paper introduced a novel traffic scene generation framework that optimizes trajectory generation through Diffusion with Transformer (DiT) blocks. The model effectively fuses latent features, historical trajectories, HD maps, and traffic signal data using transformer-based encoders with attention mechanisms. A key contribution is the multimodal trajectory decoder, which generates a wide range of future trajectories, enhancing the diversity and realism of the generated traffic scenes. Experimental results show that our approach sets a new standard for realism and diversity in traffic scene generation. Future work will focus on improving robustness for more complex urban scenarios and handling more agents.
\bibliography{ref}
\bibliographystyle{IEEEtran}

\end{document}